\documentclass[10pt,twocolumn,letterpaper]{article}

\usepackage{iccv}
\usepackage{times}
\usepackage{epsfig}
\usepackage{graphicx}
\usepackage{amsmath}
\usepackage{amssymb}
\usepackage{algorithm}
\usepackage{algorithmic}
\usepackage{bbm}
\usepackage{amsfonts}
\usepackage{mathrsfs}
\usepackage{multirow}
\usepackage{booktabs}
\usepackage{enumitem}

\usepackage[pagebackref=true,breaklinks=true,letterpaper=true,colorlinks,bookmarks=false]{hyperref}

\iccvfinalcopy 

\def\iccvPaperID{406} 

\begin{document}

\title{Disentangling Causal Substructures for Interpretable and Generalizable Drug Synergy Prediction}

\author{Yi Luo{$^{12}$},~ Haochen Zhao{$^{3}$},~ Xiao Liang{$^{12}$},~ Yiwei Liu{$^{12}$},~ Yuye Zhang{$^{12}$},~ Xinyu Li{$^{12}$},~ Jianxin Wang{$^{12*}$}
\\\vspace{-8pt}{\small~}\\
{$^{1}$}School of Computer Science and Engineering, Central South University, Changsha, 410083, China\\
{$^{2}$}Hunan Provincial Key Lab on Bioinformatics, Central South University, Changsha, 410083, China\\
{$^{3}$}Big Data Institute, Central South University, Changsha, 410083, China\\
{{$^{*}$}Corresponding to: \tt{jxwang@mail.csu.edu.cn}}
}


\maketitle
\thispagestyle{empty}

\begin{abstract}
Drug synergy prediction is a critical task in the development of effective combination therapies for complex diseases, including cancer. Although existing methods have shown promising results, they often operate as black-box predictors that rely predominantly on statistical correlations between drug characteristics and results. To address this limitation, we propose CausalDDS, a novel framework that disentangles drug molecules into causal and spurious substructures, utilizing the causal substructure representations for predicting drug synergy. By focusing on causal sub-structures, CausalDDS effectively mitigates the impact of redundant features introduced by spurious substructures, enhancing the accuracy and interpretability of the model. In addition, CausalDDS employs a conditional intervention mechanism, where interventions are conditioned on paired molecular structures, and introduces a novel optimization objective guided by the principles of sufficiency and independence. Extensive experiments demonstrate that our method outperforms baseline models, particularly in cold start and out-of-distribution settings. Besides, CausalDDS effectively identifies key substructures underlying drug synergy, providing clear insights into how drug combinations work at the molecular level. These results underscore the potential of CausalDDS as a practical tool for predicting drug synergy and facilitating drug discovery.\end{abstract}

\section{Introduction}

The prediction of drug synergy is an essential task in the field of drug discovery. When drugs are used in combination, if their combined effect exceeds the sum of the effects of each drug used individually, it is referred to as synergy\cite{jia2009mechanisms,chen2025computational,huusari2025predicting}. Compared to monotherapies, drug combinations can improve treatment efficacy, reduce toxicity, and side effects \cite {plana2022independent,meric2021enhancing,lehar2009synergistic}. For example, triple-negative breast cancer (TNBC) is a highly aggressive malignancy characterized by a high rate of metastasis and a poor prognosis. While Lapatinib or Rapamycin alone exhibits limited therapeutic efficacy, their combination has been reported to significantly enhance the apoptosis of TNBC cells\cite{jaaks2022effective,liu2011combinatorial}. Researchers often rely on high-throughput combinatorial screening to discover potential drug combinations, but the sheer number of possible combinations renders this approach time-consuming and labor-intensive.

Over recent years, numerous computational approaches have been proposed to address the drug synergy prediction task\cite{ianevski2019prediction,cheng2019network}. Among them, graph neural networks (GNNs) have recently shown great success in drug synergy prediction\cite{zhang2023kgansynergy,zhang2022predicting}. By representing a drug as a graph, i.e., considering an atom and a bond in a drug as a node and an edge in a graph, respectively. Such methods aim to capture intrinsic chemical properties or substructures and their influence on synergy. DeepDDs\cite{wang2022deepdds} is a deep learning model based on a graph neural network and attention mechanism to identify drug combinations. JointSyn\cite{li2024dual} predicts drug synergy by combining two drug structures into a single graph and utilizes dual-view jointly learning. CGMS\cite{wang2023complete} models a drug combination and a cell line as a heterogeneous complete graph, and generates the whole-graph embedding to characterize their interaction by leveraging the heterogeneous graph attention network. SDDSynergy\cite{liu2024sddsynergy} decomposes the task of predicting drug synergy by first forecasting the effects of individual substructures on cancer cell lines and emphasizes the role of crucial substructures using an attention mechanism between drugs and cell lines. Although these methods demonstrate promising performance in predicting drug synergy, they all suffer from limitations in interpretability and generalization, which constrain their applicability and reliability in real-world scenarios.

In chemical reality, the pharmacological properties of a molecule are primarily determined by key substructures\cite{wang2025token,ching2018opportunities,mcnaught1997compendium}. Here, we define these key substructures that govern the therapeutic properties of a drug as causal substructures. Substructures outside the causal substructures have minimal impact on the drug's therapeutic properties, and we define these as spurious substructures. Inevitable redundancy and noise in molecular structures can hinder the ability of graph neural networks (GNNs) to accurately learn the representations of molecular graph information\cite{lv2022causality}. Therefore, identifying and modeling these causal substructures is crucial for generating robust and generalizable drug representations\cite{yuan2021explainability,ying2019gnnexplainer}. Existing drug synergy prediction approaches are mostly based on attention or pooling. Attention-based and pooling-based approaches often lack a clear mechanism for differentiating between causal and spurious substructures in molecular graphs\cite{lee2023shift,fan2022debiasing,wu2023chemistry}. This makes them vulnerable to distributional noise, degrading performance and interpretability, especially in out-of-distribution or cold-start scenarios\cite{li2023invariant,ming2022impact}. 

To this end, we propose CausalDDS, the first deep learning framework for disentangling causal substructures from molecular graphs to predict drug synergy. CausalDDS leverages factual signals from causal substructures to derive robust and disentangled molecular representations, thereby facilitating accurate prediction of drug synergy. As shown in Fig. 1, CausalDDS disentangles the causal substructures within the molecular graph by conditioning on the paired drug in the representation space. Subsequently, we implement a conditional intervention module, parameterizing the backdoor adjustment, which encourages the model to produce robust predictions by incorporating causal substructures, paired drugs, and diverse spurious substructures. Furthermore, we design a novel loss function guided by the principles of sufficiency and independence, which enforces independence from spurious features while ensuring that the representation of causal substructure retains sufficient information for accurate prediction. Finally, we concatenate the causal substructure representations of both drugs with the cell line representation to predict the synergy. Extensive experiments have demonstrated that CausalDDS significantly enhances performance compared to state-of-the-art drug synergy prediction methods, particularly in challenging scenarios such as cold-start situations and out-of-distribution (OOD) experimental settings. CausalDDS not only achieves superior prediction accuracy under these conditions but also provides interpretable insights into the underlying mechanisms of synergy, offering valuable explanations for the model's decisions. In addition, the effectiveness of CausalDDS has been validated through several real-world examples, demonstrating its robustness and practical applicability in diverse drug discovery contexts.

In summary, our contributions are as follows:

\begin{enumerate}

    \item We propose a drug synergy prediction model based on disentangling causal substructures from drug molecular graphs. By focusing on causal substructures, the impact of redundant features introduced by spurious substructures is avoided, thereby improving prediction accuracy and interpretability.

    \item We design the optimization objection based on the sufficiency and independence principles. The sufficiency principle ensures that the causal substructures of drugs are sufficient to predict the synergy. The independence principle constrains spurious substructures to be independent of the label. Moreover, we design the causal intervention framework which can help the model learns the causality between drug causal substructures and synergy prediction, regardless of various spurious substructures and distribution shifts.

    \item We conduct extensive experiments under six experimental setups on two public datasets and the results show that CausalDDS consistently outperforms existing methods. CausalDDS also performs better under sparse data and independent test data. Interpretability analyses and case studies demonstrate the potential utility value of the model.
\end{enumerate}

\begin{figure*}[h]
    \centering
    \includegraphics[width=\linewidth]{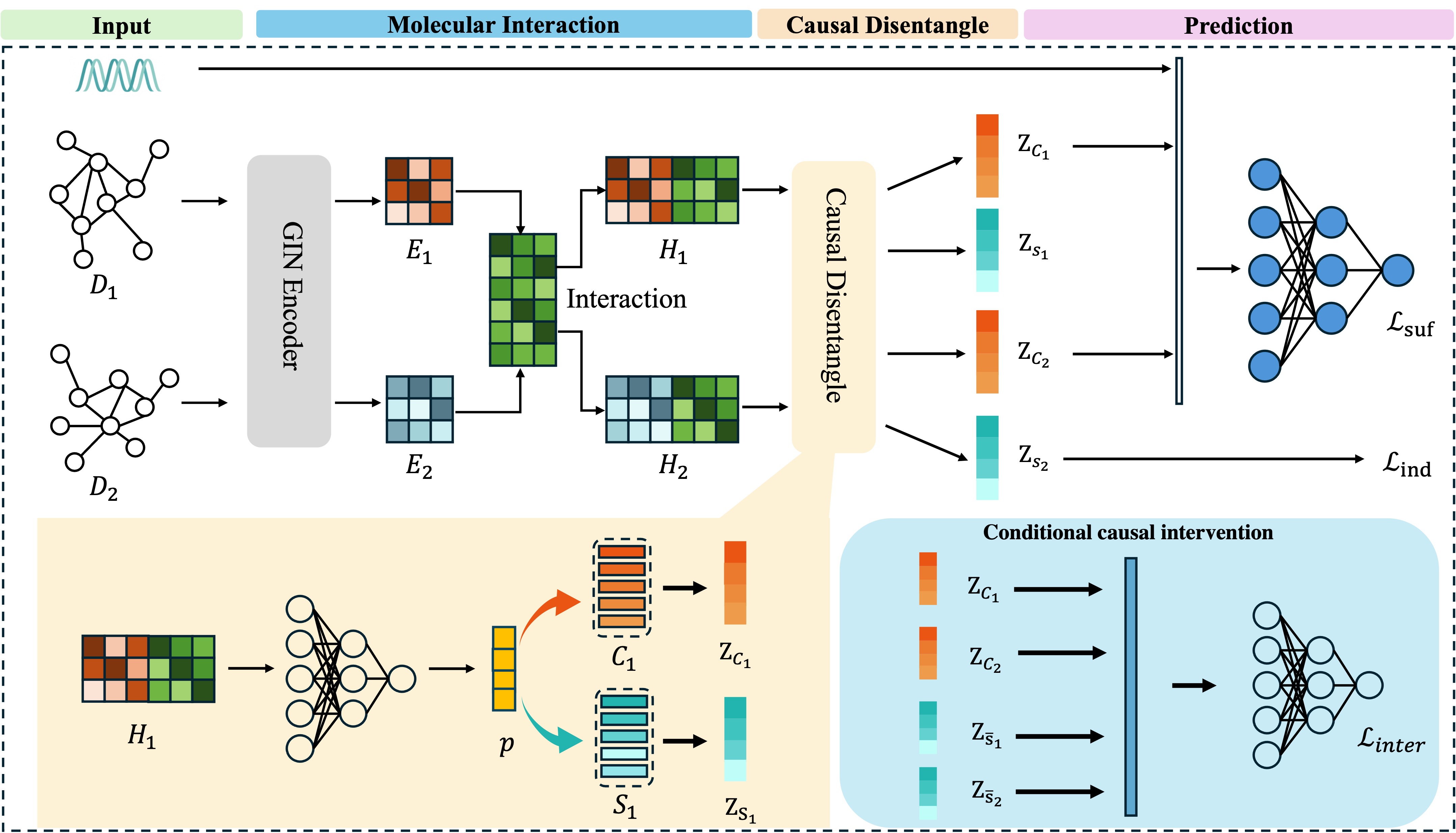}
    \caption{\textbf{Overview of the CausalDDS framework.} The inputs to the model are the SMILES string of two drugs and the gene expression data of the cell line. The output is a predicted drug synergy score or binary label. The drug molecules are first encoded using Graph Isomorphism Networks (GINs), where each row in the resulting representation corresponds to an aggregated embedding of local atomic and bond-level substructures. Then, the encoded representations of two drugs are fed into a molecular interaction module to capture pairwise atom-level interactions and derive interaction-aware molecular representations. Thirdly, the causal disentangle module uses an MLP to estimate an atom-level importance score from the embedding matrix. Based on the importance score, we separate the causal substructure and the spurious substructure by masking the molecular representation. Finally, the causal substructure representations of the selected drug pair, together with the gene expression data of the target cell line, are jointly input into a fully connected neural network to predict drug synergy. The conditional causal intervention module is given a drug pair; spurious substructures are extracted by modeling their interactions with other drugs in the training set, and are then used to optimize the model through conditional intervention. Optimization objective defined according to the principles of sufficiency and independence. The final loss function is defined as the sum of three losses.}
    \label{fig:matched_molecules}
\end{figure*}

\section{Results}

\subsection{CausalDDS framework.~}
The proposed CausalDDS framework is shown in Fig. 1.  CausalDDS consists of five steps. (1) Input. The input of CausalDDS is the SMILES\cite{weininger1988smiles} strings of drug compounds and the gene expression data of cell lines. (2) Molecular interaction. This module encodes molecular graphs into vector representations and integrates the interaction between two drugs into their representations. (3) Causal disentangle module. This module separates the causal substructure and the spurious substructure from the molecular graph of the drug. (4) Conditional Causal Intervention Module. This module alleviates the confounding effect of spurious substructure by backdoor adjustment. (5) Prediction module. The representations of causal substructure in two drugs and the cell line representation are concatenated and processed by a three-layer fully connected neural network with dropout to predict the drug synergy.

\subsection{CausalDDS outperforms existing methods on six experimental settings}

\begin{table*}[ht]
\centering
\caption{Classification comparison results of CausalDDS and baselines on the DrugCombDB dataset.}
\scriptsize
\begin{tabular}{p{2.5cm}p{2.5cm}p{2.5cm}p{2.5cm}p{2.5cm}p{2.5cm}}
\toprule
\textbf{Settings} & \textbf{Methods} & 
{AUC} & {AUPR} & {KAPPA} & {F1} \\
\midrule

\multirow{7}{*}{warm start}
& DeepDDs &0.858(0.011)  &0.377(0.028)  &0.375(0.020)  &0.404(0.020)   \\
& HypergraphSynergy &0.848(0.010)  &0.433(0.022) &0.281(0.036)  &0.476(0.022)   \\
& MDNNSyn &0.882(0.015)  &0.426(0.047)  &0.270(0.050)  &0.457(0.034)   \\
& AttenSyn &0.881(0.005)  &0.430(0.027)  &0.411(0.017)  &0.437(0.017)   \\
& SDDSynergy &0.849(0.012)  &0.341(0.022)  &0.265(0.055)  &0.286(0.058)   \\
& JointSyn &\underline{0.888(0.008)}  &\underline{0.491(0.021)}  &\underline{0.478(0.021)}  &\underline{0.502(0.021)}   \\
& CausalDDS & \textbf{0.917(0.004)} & \textbf{0.542(0.025)} & \textbf{0.512(0.014)} & \textbf{0.537(0.013)}  \\
\midrule

\multirow{7}{*}{unseen drug pair}
& DeepDDs &\underline{0.835(0.017)}&0.352(0.045)&0.342(0.043)&0.366(0.044)\\
& HypergraphSynergy &0.807(0.018)&0.294(0.047)&0.220(0.030)&0.388(0.029)\\
& MDNNSyn &0.851(0.016)&0.351(0.041)&0.215(0.034)&\underline{0.409(0.022)}\\
& AttenSyn &0.824(0.022)&0.305(0.043)&0.307(0.044)&0.338(0.047)\\
& SDDSynergy &0.778(0.046)&0.235(0.055)&0.138(0.056)&0.153(0.057)\\
& JointSyn &0.834(0.020)&\underline{0.366(0.061)}&\underline{0.372(0.051)}&0.399(0.051)\\
& CausalDDS & \textbf{0.877(0.013)} & \textbf{0.416(0.034)} & \textbf{0.423(0.027)} & \textbf{0.454(0.027)} \\
\midrule

\multirow{7}{*}{unseen cell line}\
& DeepDDs &0.752(0.055)&0.220(0.057)&0.257(0.069)&0.291(0.066)\\
& HypergraphSynergy &0.717(0.078)&0.206(0.029)&0.115(0.061)&0.278(0.019)\\
& MDNNSyn &0.728(0.060)&0.238(0.036)&0.167(0.046)&0.318(0.028)\\
& AttenSyn &0.792(0.022)&\underline{0.279(0.043)}&0.299(0.044)&0.330(0.043)\\
& SDDSynergy &\underline{0.797(0.013)}&0.268(0.040)&0.217(0.043)&0.238(0.045)\\
& JointSyn &0.794(0.019)&0.279(0.046)&\underline{0.305(0.057)}&\underline{0.339(0.054)}\\
& CausalDDS & \textbf{0.829(0.012)} & \textbf{0.330(0.044)} & \textbf{0.365(0.034)} & \textbf{0.400(0.029)} \\
\midrule

\multirow{7}{*}{unseen drug}
& DeepDDs &0.694(0.061)&0.170(0.053)&0.166(0.106)&0.191(0.120)\\
& HypergraphSynergy &0.635(0.062)&0.171(0.063)&0.005(0.009)&0.216(0.064)\\
& MDNNSyn &0.746(0.035)&0.188(0.050)&0.136(0.051)&0.267(0.045)\\
& AttenSyn &0.728(0.028)&0.174(0.049)&0.192(0.049)&0.227(0.050)\\
& SDDSynergy &0.719(0.049)&0.149(0.055)&0.086(0.035)&0.104(0.039)\\
& JointSyn &\underline{0.751(0.040)}&\underline{0.225(0.055)}&\underline{0.254(0.072)}&\underline{0.284(0.074)}\\
& CausalDDS & \textbf{0.792(0.017)} & \textbf{0.240(0.063)} & \textbf{0.271(0.053)} & \textbf{0.323(0.050)} \\

\midrule

\multirow{7}{*}{Scaffold split}
& DeepDDs &0.834(0.017)&0.345(0.027)&0.345(0.013)&0.371(0.012)\\
& HypergraphSynergy &0.799(0.027)&0.268(0.044)&0.201(0.037)&0.376(0.034)\\
& MDNNSyn &\underline{0.848(0.018)}&0.349(0.051)&0.199(0.036)&0.402(0.036)\\
& AttenSyn &0.832(0.009)&0.315(0.031)&0.327(0.024)&0.356(0.026)\\
& SDDSynergy &0.801(0.028)&0.256(0.055)&0.172(0.103)&0.187(0.109)\\
& JointSyn &0.844(0.009)&\underline{0.377(0.037)}&\underline{0.384(0.028)}&\underline{0.411(0.030)}\\
& CausalDDS &\textbf{0.871(0.016)}&\textbf{0.450(0.043)}&\textbf{0.435(0.020)}&\textbf{0.464(0.029)}\\

\midrule
\multirow{7}{*}{SIMPD split}
& DeepDDs &0.867(0.011)&0.424(0.014)&0.413(0.010)&0.439(0.011)\\
& HypergraphSynergy &0.849(0.003)&0.439(0.018)&0.266(0.025)&0.478(0.015)\\
& MDNNSyn &0.888(0.009)&0.429(0.028)&0.279(0.012)&0.463(0.018)\\
& AttenSyn &0.882(0.007)&0.428(0.011)&0.419(0.009)&0.444(0.009)\\
& SDDSynergy &0.838(0.020)&0.324(0.030)&0.247(0.066)&0.266(0.070)\\
& JointSyn &\underline{0.894(0.009)}&\underline{0.493(0.017)}&\underline{0.473(0.016)}&\underline{0.497(0.016)}\\
& CausalDDS &\textbf{0.914(0.004)}&\textbf{0.520(0.013)}&\textbf{0.490(0.007)}&\textbf{0.517(0.009)}\\

\bottomrule
\end{tabular}
\footnotesize \raggedright The best results for each metric are highlighted in bold, and the second-best results are underlined.
\label{tab:grouped_results}
\end{table*}

\begin{table*}[ht]
\centering
\caption{Classification comparison results of CausalDDS and baselines on the ONEIL-COSMIC dataset.}
\scriptsize
\begin{tabular}{p{2.5cm}p{2.5cm}p{2.5cm}p{2.5cm}p{2.5cm}p{2.5cm}}
\toprule
\textbf{Settings} & \textbf{Methods} & 
{AUC} & {AUPR} & {KAPPA} & {F1} \\
\midrule

\multirow{7}{*}{warm start}
& DeepDDs &0.846(0.005)&0.760(0.004)&0.508(0.009)&0.675(0.005)\\
& HypergraphSynergy &0.863(0.003)&0.800(0.005)&0.560(0.022)&\underline{0.718(0.009)}\\
& MDNNSyn &0.845(0.016)&0.757(0.024)&0.499(0.029)&0.696(0.016)\\
& AttenSyn &0.840(0.005)&0.748(0.009)&0.505(0.015)&0.674(0.013)\\
& SDDSynergy &0.829(0.016)&0.729(0.029)&0.478(0.055)&0.645(0.055)\\
& JointSyn &\underline{0.869(0.004)}&\underline{0.808(0.006)}&\underline{0.568(0.010)}&0.712(0.006)\\
& CausalDDS & \textbf{0.887(0.005)} & \textbf{0.820(0.009)} & \textbf{0.582(0.018)} & \textbf{0.737(0.009)}  \\
\midrule

\multirow{7}{*}{unseen drug pair}
& DeepDDs &0.800(0.010)&0.683(0.033)&0.439(0.019)&0.630(0.024)\\
& HypergraphSynergy &0.812(0.016)&\underline{0.710(0.030)}&\underline{0.458(0.031)}&\underline{0.663(0.021)}\\
& MDNNSyn &0.809(0.022)&0.702(0.031)&0.457(0.033)&0.661(0.020)\\
& AttenSyn &0.773(0.023)&0.654(0.042)&0.390(0.038)&0.604(0.023)\\
& SDDSynergy &0.760(0.061)&0.635(0.089)&0.377(0.097)&0.591(0.084)\\
& JointSyn &\underline{0.814(0.014)}&0.706(0.025)&0.457(0.027)&0.663(0.021)\\
& CausalDDS & \textbf{0.824(0.017)} & \textbf{0.724(0.047)} & \textbf{0.470(0.043)} & \textbf{0.674(0.026)} \\
\midrule

\multirow{7}{*}{unseen cell line}\
& DeepDDs &0.797(0.011)&0.682(0.019)&0.431(0.011)&0.625(0.005)\\
& HypergraphSynergy &\underline{0.808(0.007)}&\underline{0.702(0.009)}&\underline{0.453(0.017)}&\underline{0.663(0.008)}\\
& MDNNSyn &0.775(0.009)&0.657(0.027)&0.401(0.028)&0.631(0.014)\\
& AttenSyn &0.790(0.012)&0.669(0.018)&0.415(0.029)&0.610(0.034)\\
& SDDSynergy &0.775(0.011)&0.643(0.034)&0.396(0.029)&0.593(0.033)\\
& JointSyn &0.770(0.013)&0.646(0.023)&0.378(0.034)&0.555(0.037)\\
& CausalDDS & \textbf{0.820(0.006)} & \textbf{0.714(0.015)} & \textbf{0.467(0.004)} & \textbf{0.674(0.004)} \\
\midrule

\multirow{7}{*}{unseen drug}
& DeepDDs &0.651(0.033)&0.501(0.039)&0.205(0.035)&0.460(0.030)\\
& HypergraphSynergy &0.676(0.018)&0.537(0.039)&0.249(0.033)&\underline{0.556(0.015)}\\
& MDNNSyn &0.670(0.017)&0.510(0.024)&0.227(0.034)&0.556(0.016)\\
& AttenSyn &0.662(0.015)&0.515(0.030)&0.217(0.026)&0.495(0.028)\\
& SDDSynergy &0.647(0.053)&0.476(0.063)&0.180(0.062)&0.440(0.019)\\
& JointSyn &\underline{0.696(0.018)}&\underline{0.567(0.023)}&\underline{0.264(0.033)}&0.492(0.034)\\
& CausalDDS & \textbf{0.729(0.027)} & \textbf{0.605(0.042)} & \textbf{0.285(0.049)} & \textbf{0.589(0.022)} \\
\midrule

\multirow{7}{*}{Scaffold split}
& DeepDDs &0.794(0.014)&0.680(0.019)&0.419(0.017)&0.617(0.016)\\
& HypergraphSynergy &0.816(0.009)&\underline{0.719(0.018)}&0.444(0.024)&\underline{0.663(0.014)}\\
& MDNNSyn &0.803(0.014)&0.705(0.017)&0.440(0.025)&0.650(0.015)\\
& AttenSyn &0.777(0.024)&0.657(0.036)&0.388(0.045)&0.599(0.043)\\
& SDDSynergy &0.776(0.035)&0.643(0.055)&0.382(0.043)&0.572(0.037)\\
& JointSyn &\underline{0.820(0.010)}&0.717(0.020)&\underline{0.469(0.018)}&0.657(0.012)\\
& CausalDDS &\textbf{0.824(0.007)}&\textbf{0.724(0.018)}&\textbf{0.476(0.013)}&\textbf{0.669(0.009)} \\

\midrule
\multirow{7}{*}{SIMPD split}
& DeepDDs &0.842(0.009)&0.753(0.014)&0.507(0.018)&0.673(0.013)\\
& HypergraphSynergy &0.874(0.004)&0.800(0.010)&0.555(0.016)&\underline{0.726(0.007)}\\
& MDNNSyn &0.854(0.017)&0.768(0.025)&0.522(0.030)&0.703(0.020)\\
& AttenSyn &0.842(0.010)&0.757(0.013)&0.519(0.022)&0.681(0.012)\\
& SDDSynergy &0.871(0.005)&0.794(0.011)&0.559(0.007)&0.702(0.007)\\
& JointSyn &\underline{0.876(0.006)}&\underline{0.804(0.013)}&\underline{0.577(0.019)}&0.721(0.014)\\
& CausalDDS &\textbf{0.884(0.004)}&\textbf{0.815(0.009)}&\textbf{0.585(0.019)}&\textbf{0.738(0.008)}\\

\bottomrule
\end{tabular}
\footnotesize \raggedright The best results for each metric are highlighted in bold, and the second-best results are underlined.
\label{tab:grouped_results}
\end{table*}

We benchmark CausalDDS against state-of-the-art approaches across two drug synergy prediction tasks: classification and regression. For each task, model performance is quantified using four evaluation metrics under four experimentally realistic scenarios. These experiments are conducted on two widely used public datasets, DrugCombDB\cite{liu2020drugcombdb} and ONEIL-COSMIC\cite{o2016unbiased}, thereby ensuring both the robustness and generalizability of the comparative analysis. To better evaluate the prediction performance of our model and the baseline in various application scenarios, we set up six different experimental settings: warm start, unseen drug pair, unseen cell line, unseen drug, scaffold split, and SIMPD split. A detailed description of the six experimental settings is provided in the Methods section. For the classification tasks, we binary the synergy scores according to a threshold provided by previous studies\cite{rafiei2023deeptrasynergy,wang2023granularity} to generate synergistic (positive) and non-synergistic (negative) samples. A sample with a measured synergy score higher than 10 is classified as a positive sample; otherwise, it is classified as a negative sample. For classification results, we report the area under the receiver operating characteristic (AUC), the area under the precision-recall curve (AUPR), the Cohen's Kappa (KAPPA), and the F1 score (F1). For regression results, we report the Root Mean Square Error (RMSE), the Mean Absolute Error (MAE), the Pearson Correlation Coefficient (PCC), and the R-Squared (R\textsuperscript{2}). These metrics and datasets are widely used in related works\cite{liu2025building,wu2023hybrid}. For benchmarking, we compare our method with six state-of-the-art approaches: HypergraphSynergy\cite{liu2022multi}, MDNNSyn\cite{li2024mdnnsyn}, DeepDDS\cite{wang2022deepdds}, AttenSyn\cite{wang2023attensyn}, SDDSynergy\cite{liu2024sddsynergy}, and JointSyn\cite{li2024dual} for classification, and HypergraphSynergy, MDNNSyn, DeepDDS, SynergyX\cite{guo2024synergyx}, SDDSynergy, and JointSyn for regression. Details of the baselines can be found in Supplementary Note 1.

The classification results on the DrugCombDB dataset under four evaluation protocols are reported in Table 1. Overall, our method achieves superior performance over contrastive methods across all six scenarios, demonstrating both high predictive accuracy and strong generalization capabilities. In the warm-start scenario, our approach delivers the highest scores in all evaluation metrics. Specifically, it achieves an AUC of 0.917 ± 0.004, AUPR of 0.542 ± 0.025, Kappa of 0.512 ± 0.014, and F1-score of 0.537 ± 0.013. Compared to the second-best baseline in each metric, these correspond to relative improvements of 3.3\%, 10.4\%, 7.1\%, and 7.0\%, respectively. We conducted the t-test and the result shows that the statistical significance of these improvement in four metrics ($p<0.05$). In the scenario of an unseen drug pair, CausalDDS achieves relative improvements in KAPPA (0.423 ± 0.027) with 13.2\% and in F1 (0.454 ± 0.027) with 11.0\%, compared with the second-best baselines JointSyn (0.372 ± 0.051) and MDNNSyn (0.409 ± 0.022) with $p<0.05$. These enhancements are statistically significant ($p<0.05$). In the unseen cell line scenario, our method achieves an AUC of 0.8289 ±0.012, outperforming the second-best baseline by 4.3\%. The advantage is even more pronounced in AUPR, Kappa, and F1, where relative improvements exceed 18\%. In the unseen drug scenario, CausalDDS achieves the best predictive performance with an AUC of 0.792 ± 0.017 and AUPR of 0.240 ± 0.064. For the scaffold split scenario, our method improved the average AUPR and F1 scores by 19.4\% and 12.9\% over the second-best method (t-test, $p<0.05$). CausalDDS achieves the best performance across all evaluation metrics in the SIMPD split, with the highest AUC of 0.914 ± 0.004. This demonstrates significant enhancements versus other models under more realistic generalization testing.

\begin{figure*}[h]
    \centering
    \includegraphics[width=\linewidth]{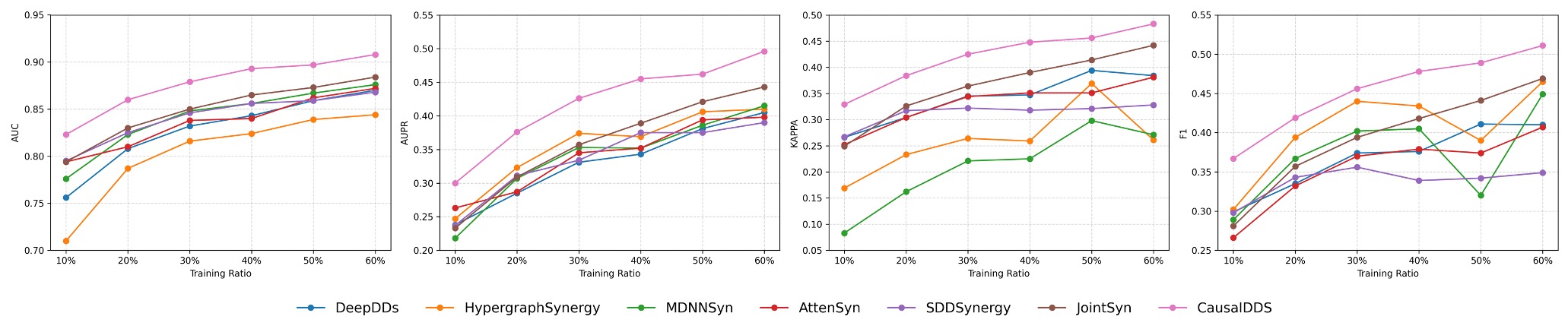}
    \caption{\textbf{Performance evaluation with scarce data on the DrugCombDB dataset}}
    \label{fig:matched_molecules}
\end{figure*}

The classification performance comparisons on the ONEIL-COSMIC dataset under four experimental settings are provided in Table 2. In the six experimental settings on the ONEIL-COSMIC dataset, CausalDDS consistently demonstrates strong performance across all metrics (AUC, AUPR, Kappa, F1). In the scenario of warm start, CausalDDS achieves the highest performance with an AUC of 0.887 ± 0.005, an AUPR of 0.820 ± 0.009, a Kappa of 0.582 ± 0.018, and an F1-score of 0.737 ± 0.009, showing relative improvements of 2.1\%, 1.5\%, 2.5\%, and 2.6\%, respectively, compared to the second-best baseline. In the scenario of unseen drug pair, CausalDDS statistically significant again outperforms the other baselines, achieving an AUC of 0.824 ± 0.017 and a Kappa of 0.470 ± 0.043, with an F1-score of 0.674 ± 0.026. In the scenario of unseen cell line, CausalDDS achieves an average AUC of 0.820 ± 0.006 and AUPR of 0.714 ± 0.015, compared with the second-best HypergraphSynergy’s AUC of 0.808 ± 0.007 and AUPR of 0.702 ± 0.009. Finally, in the scenario of unseen drug, CausalDDS outperforms the second-best method, HypergraphSynergy, by 8.0\% and 5.9\% in KAPPA and F1 at $p<0.05$, respectively. Specifically, CausalDDS achieves a KAPPA of 0.285 ± 0.049 and an F1 of 0.589 ± 0.022. CausalDDS also achieves the best performance in the scaffold split scenario, with an AUC of 0.824 ± 0.007, AUPR of 0.724 ± 0.018, Kappa of 0.476 ± 0.013, and F1 of 0.669 ± 0.009. These results surpass the strongest baseline across all four metrics. Overall, CausalDDS shows robust and consistent performance across various experimental settings, demonstrating its ability to generalize well and outperform competing methods.

The regression performance results on the DrugCombDB and ONEIL-COSMIC datasets are summarized in Supplementary Tables 1 and 2. CausalDDS achieves excellent results on both datasets, although in some experimental settings, some metrics are slightly lower than the best-performing methods. These results demonstrate that CausalDDS is a highly competitive model for drug synergy regression tasks.

\subsection{Performance evaluation with independent test data and scarce data}

To better assess its performance in real-world applications, we trained the models on the DrugCombDB and the ONEIL-COSMIC, and tested them on the AstraZeneca dataset\cite{menden2019community} and the ALMANAC dataset\cite{holbeck2017national}. As shown in Fig. 3 and Supplementary Tables 3-6, CasualDDS achieves optimal results across the majority of metrics and conditions. These results indicate that our model has stronger potential for real-world applications compared to existing approaches. 

In practical applications, especially when dealing with new drugs or rare diseases, obtaining large amounts of labeled data may not be possible. By simulating scenarios with limited training data, we can evaluate whether the model is able to extract useful patterns from a small number of training samples. To evaluate the performance of CausalDDS and the baselines under sparse conditions, we train each model on 10\%, 20\%, 30\%, 40\%, 50\%, and 60\% of the datasets and test on the remaining 90\%, 80\%, 70\%, 60\%, 50\%, and 40\% of the datasets, respectively. To fairly compare, we perform a 5-times Monte Carlo cross-validation on two datasets. The model performance comparisons on the DrugCombDB dataset under different sparse conditions are presented in Fig. 2 (Supplementary Tables 7 and 8). CausalDDS consistently outperforms competing methods across all evaluation metrics (t-test, $p<0.05$. Its performance steadily improves with larger training ratios, highlighting the robustness and scalability of the model under increasing data availability. In contrast, other methods show either slower gains or noticeable fluctuations, underscoring the advantage of CausalDDS’s stable upward trend. On the O’Neil-COSMIC dataset (Supplementary Fig. 1), although the smaller scale reduces the performance gap among methods, CausalDDS still achieves competitive results. Meanwhile, MDNNSyn exhibits larger fluctuations, further emphasizing the robustness of CausalDDS under more challenging conditions. Together, these results demonstrate that CausalDDS maintains reliable performance across datasets of different scales, consistently delivering steady and effective improvements. Such steady improvement is particularly advantageous in practical scenarios, where additional training data can be expected to further enhance predictive performance without compromising stability.

\begin{figure}[h]
    \centering
    \includegraphics[width=\linewidth]{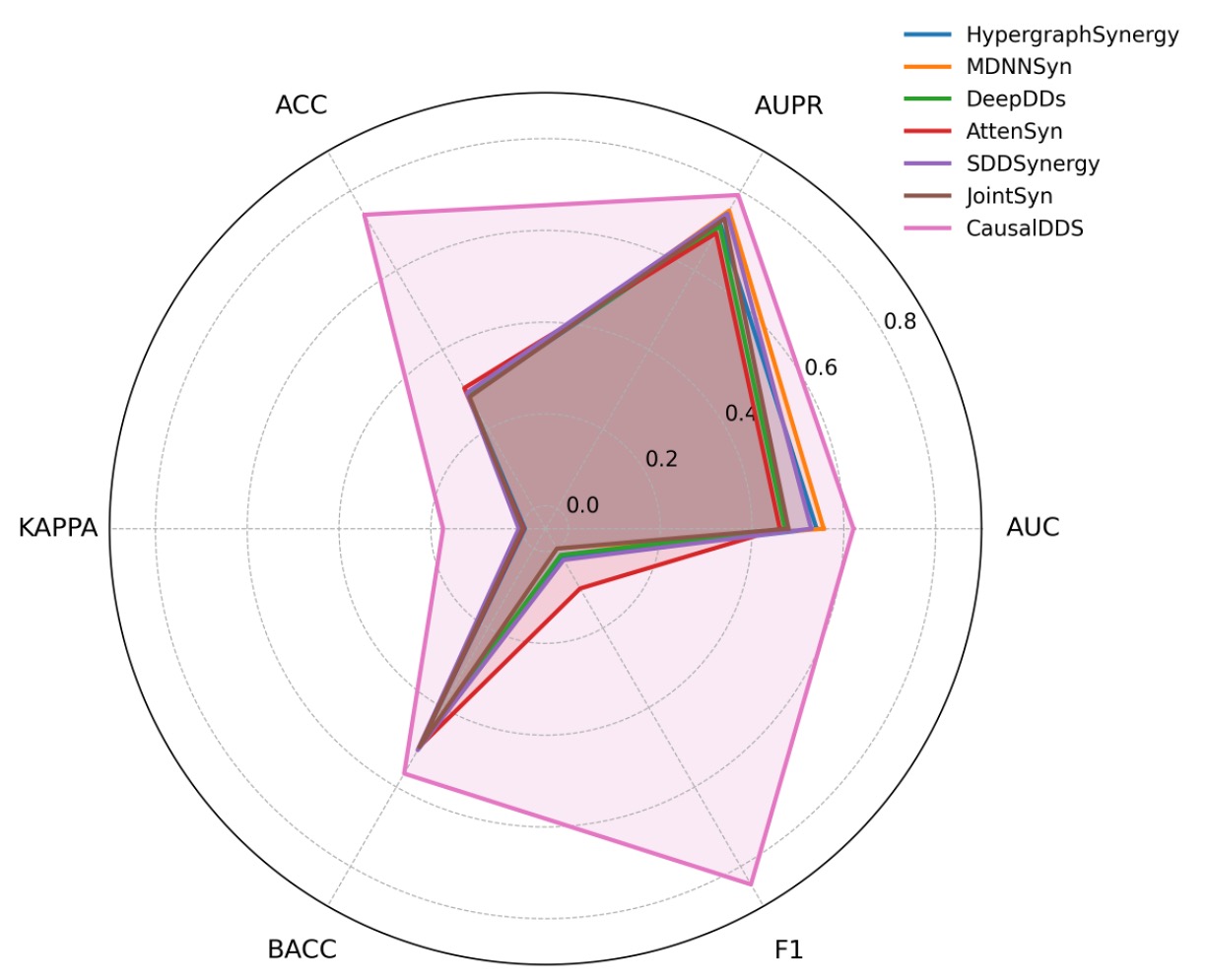}
    \caption{\textbf{Performance evaluation on the AstraZeneca Dataset using models trained on the DrugCombDB Dataset}}
    \label{fig:matched_molecules}
\end{figure}

\subsection{Impact of each module of CausalDDS on predictive performance}

Here, we conduct an ablation study to examine the impact of individual modules on CausalDDS. The configurations of three different variants of CausalDDS are described as follows. WODisentangle directly concatenates drug molecular graph representations obtained via GIN with cell line gene expression features, which are then fed into a multilayer perceptron (MLP) for prediction. WOSufficiency is a variant model that removes the sufficiency-based loss from the final optimization objective, while WOIndependence removes the independence-based loss. WOIntervention excludes the conditional causal intervention module.

The results under four scenarios on the ONEIL-COSMIC dataset are shown in Supplementary Fig. 2 and Supplementary Table 9. For WODisentangle, we observe a significant decrease in all four scenarios, demonstrating that the incorporation of causal disentanglement into molecular graph models enhances the ability to capture underlying causal relationships between drug molecules and biological events, leading to improved model generalization capability. We can see that WOIndependence and WOSufficiency perform slightly worse than CausalDDS, indicating the necessity of two optimization losses with causal disentangle in our model. Additionally, WOIntervention yields a lower performance without the causal intervention module, which mitigates spurious correlations and encourages the model to focus on causal substructure.

\subsection{Interpretability with causal substructure}

\begin{figure*}
    \centering
    \includegraphics[width=0.9\linewidth]{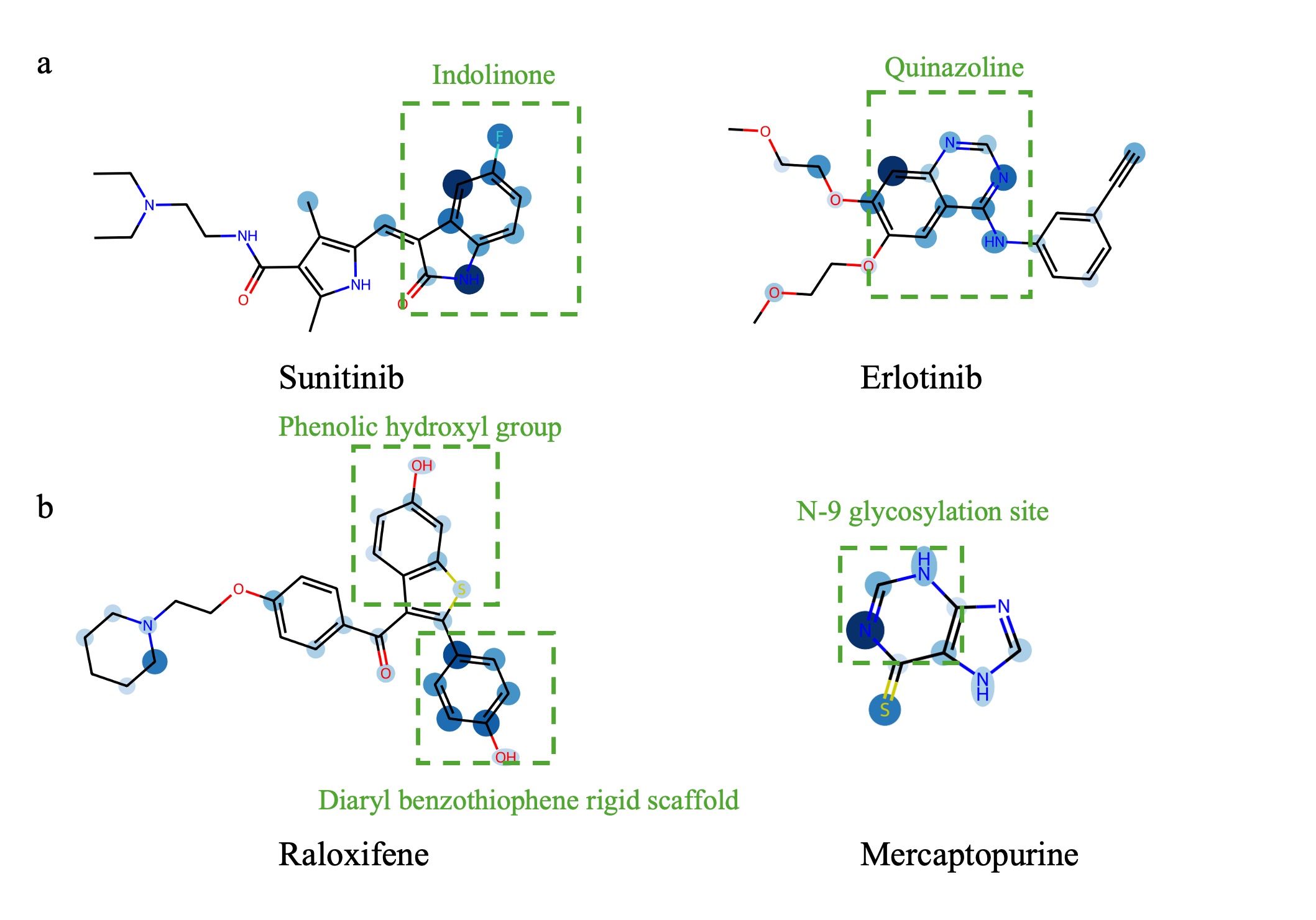}
    \caption{\textbf{Interpretability analysis of CausalDDS for explaining drug synergy mechanisms. a,} Visualization of the causal substructures underlying the synergistic drug combination of sunitinib and erlotinib in the COLO320 cell line. The substructure indolinone for sunitinib and the substructure quinazoline for erlotinib, playing decisive roles, are highlighted with green dashed boxes. \textbf{b,} Visualization of the causal substructures underlying the synergistic drug combination of raloxifene and mercaptopurine in the SK-MEL-28 cell line. The substructure phenolic hydroxyl group and diaryl benzothiophene rigid scaffold for raloxifene and the substructure N-9 glycosylation site for Mercaptopurine are highlighted with green dashed boxes.}
    \label{fig:placeholder}
\end{figure*}

It is well established that the causal substructures of drug molecules play a decisive role in determining their pharmacological properties. Identifying these substructures is critical for understanding drug synergy, facilitating drug repurposing, and guiding rational drug design. 

Sunitinib is a receptor tyrosine kinase inhibitor approved for the treatment of renal cell carcinoma (RCC) and imatinib-resistant gastrointestinal stromal tumors (GIST). Erlotinib, an EGFR tyrosine kinase inhibitor, is used in the management of certain Non-Small-Cell Lung Cancers (NSCLC) and advanced pancreatic cancers\cite{christensen2007preclinical,gan2009sunitinib,blay2009advanced,quek2009gastrointestinal}. The COLO320 colorectal cancer cell line exhibits poor response to monotherapy with erlotinib. However, when administered in combination with sunitinib, a marked synergistic anticancer effect is observed\cite{bozec2009combination,li2016preclinical,nukatsuka2012combination}. To elucidate the molecular basis of this synergy, we apply CausalDDS to visualize the causal substructures that contribute to the combinatorial efficacy of sunitinib and erlotinib in COLO320 cells (Fig. 4a). The nodes are colored according to their scores assigned by the model, with darker colors indicating higher values. Green boxes highlight causal substructures identified as critical for synergy. Erlotinib contains a quinazoline, critical for EGFR inhibition. However, since COLO320 cells are not primarily dependent on EGFR signaling, erlotinib alone is insufficient to produce significant therapeutic effects\cite{sigismund2018emerging,hampton2020new}. Sunitinib has an indolinone scaffold that targets multiple receptor tyrosine kinases, including VEGFR and PDGFR\cite{giuliano2015resistance}. Therefore, the combination of erlotinib and sunitinib achieves a broader blockade of compensatory signaling pathways, resulting in enhanced tumor suppression through a synergistic mechanism\cite{pan2011synergistic,zhou2021resistance}. 

The SK-MEL-28 melanoma cell line is a widely used experimental model to investigate melanoma biology and evaluate novel therapeutic strategies\cite{sambade2011melanoma}. According to the NCI-ALMANAC dataset, the combination of raloxifene and mercaptopurine shows notable efficacy in this cell line\cite{holbeck2017national,sidorov2019predicting}. From a mechanistic perspective, the phenolic hydroxyl group of Raloxifene constitutes a critical determinant for recognition and metabolic modification of the transporter, while its rigid sulfur-containing scaffold provides electronic and hydrophobic interactions that strengthen the affinity of the receptor and transporter\cite{yang2021g,vaughan2013mechanisms}. In contrast, the N-9 glycosylation site of mercaptopurine is indispensable for its biotransformation into thioguanine nucleotides, which underpins its established antimetabolite activity\cite{esmail2021advances}. Functionally, raloxifene predominantly modulates drug efflux and metabolic environment, while activation through the N-9 site of mercaptopurine represents the essential pharmacological trigger\cite{zaza2005gene}. When used together, they result in a potential synergistic therapeutic effect. 

We also conducted analyses on the synergy mechanisms of other drug combinations, with the results presented in Supplementary Figs. 3-5. CausalDDS demonstrates not only an accurate prediction of synergistic drug combinations, but also the potential to identify key substructures responsible for synergistic effects.

\subsection{Structural optimization for drug synergy with CausalDDS.~}

In the field of drug synergy, even subtle structural variations in similar compounds can lead to significant differences in their synergistic effects. This phenomenon underscores the complexity of drug interactions and the nuanced influence of molecular structure on biological activity. Traditional drug design typically focuses on large-scale structural optimization, often lacking the precision needed to capture minor variations. As a solution, CausalDDS offers a fast and effective method for computing and testing whether modifications to the molecular structure of a drug can result in synergistic effects. 

Lapatinib and afatinib are structurally similar small-molecule tyrosine kinase inhibitors (TKIs) targeting the HER family of proteins, including HER2 and EGFR\cite{geyer2006lapatinib,opdam2012lapatinib}. Despite their structural similarity and shared mechanism of inhibition of HER-family signaling, they differ slightly in their binding modes, target affinities, and clinical uses\cite{park2016afatinib,roskoski2014erbb}. 
As shown in Fig. 5a, the yellow highlight is the largest common substructure of the two drugs. To verify the capability of CausalDDS in capturing the impact of fine-grained variations in drug substructures on combination therapy outcomes, we collected 39 drug pairs involving lapatinib tested in the EFM-192A cell line from the ONEIL-COSMIC dataset and replaced lapatinib with afatinib. It is noteworthy that afatinib was not included in the training dataset. Subsequently, we use CausalDDS to predict the synergy score for each drug pair.

After a non-exhaustive literature search, we find nine predicted drug combinations that are consistent with observations in previous studies or in clinical trials. For example, when lapatinib is replaced with afatinib, the synergy score combined with geldanamycin changed dramatically from -14.111, as recorded in the original dataset, to 9.285. Geldanamycin and afatinib have been investigated for their synergistic effects in the EFM-192A cell line, a HER2-positive breast cancer model\cite{wang2007geldanamycin,song2022synergistic}. Their combination targets both protein folding and HER2 signaling, potentially overcoming cancer cell resistance and enhancing therapeutic efficacy. Afatinib provides immediate functional inhibition of HER2's signaling activity. Geldanamycin delivers a long-term, destructive attack by targeting and eliminating the HER2 protein itself\cite{gunzer2016phase}.

We also observed that when combined with mk-4541, the synergy score changed from 19.213 to -5.850. Preclinical studies have reported that the combination of the afatinib nation with the FGFR inhibitor may be anti-synergistic (antagonistic) in certain contexts. This antagonism has been attributed to resistance mechanisms to afatinib\cite{mao2023statistical,lai2018crosstalk}. In several models, concurrent blockade of EGFR / HER2 and FGFR appears to engage compensatory, intersecting survival pathways, thus reducing the efficacy of either agent\cite{chen2013synergistic}.

We replicated this experiment with erlotinib and gefitinib, and the results are presented in Supplementary Table 11. Among the nine predicted drug combinations, eight are validated. Our experiments demonstrate that the model can capture subtle structural variations that influence drug synergy scores, thereby facilitating a faster confirmation of therapeutic efficacy during the drug design process.

\begin{figure*}
    \centering
    \includegraphics[width=0.9\linewidth]{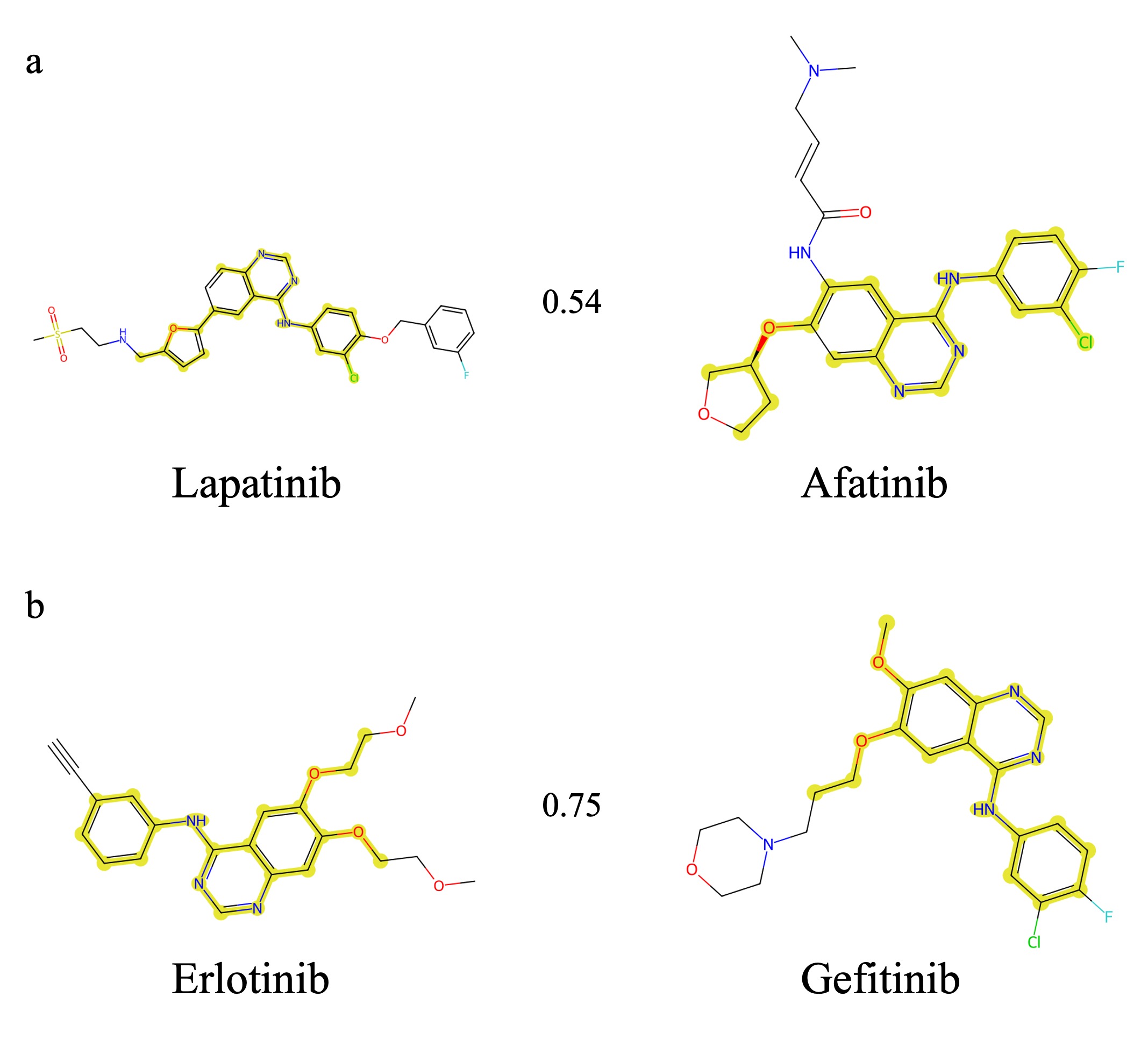}
    \caption{\textbf{The case study of structural optimization for drug synergy with CausalDDS. a,} The left side shows the 2D structure of lapatinib, and the right side shows the 2D structure of afatinib. The yellow-highlighted part is the maximum common structure of the two drugs found by the Rdkit. The score is the Tanimoto similarity. \textbf{b,} The left side shows the 2D structure of erlotinib, and the right side shows the 2D structure of gefitinib.}
    \label{fig:placeholder}
\end{figure*}

\begin{table*}[htbp]
\centering
\caption{The change in synergy score after replacing Lapatinib with Afatinib.}
\label{tab:supplementary8}
\begin{tabular}{ccccccc}
\toprule
Drug A Name  & Raw Drug Name  & Change Drug Name  &Cell line & Raw Score & Changed Score & Validation \\
\midrule
Adavosertib &Lapatinib & Afatinib & EFM-192A & 22.886 & 10.604& True \\
carboplatin  &Lapatinib & Afatinib& EFM-192A & 3.466 & -4.667 & False \\
Dexamethasone  &Lapatinib & Afatinib& EFM-192A & -43.359 & -1.722 & True \\
Etoposide  &Lapatinib & Afatinib& EFM-192A & -1.479 & 27.651 & True \\
Dasatinib  &Lapatinib & Afatinib& EFM-192A & 66.625 & 17.137 & True \\
Dinaciclib  &Lapatinib & Afatinib& EFM-192A & 16.798 & -6.197 & False \\
Geldanamycin &Lapatinib & Afatinib & EFM-192A & -14.111 & 9.285 & True \\
Niraparib &Lapatinib & Afatinib & EFM-192A & 19.700 & 1.499 & True \\
Ridaforolimus  &Lapatinib & Afatinib& EFM-192A & 52.340 & 24.550 & True \\
SN-38  &Lapatinib & Afatinib& EFM-192A & 22.561 & 0.261 & True \\
MK-4541 &Lapatinib & Afatinib & EFM-192A & 19.213 & -5.840 & True \\
\bottomrule
\end{tabular}
\end{table*}

\section{Discussion.~}
Understanding the molecular determinants of drug synergy remains a key challenge in designing effective combination therapies. In this study, we propose CausalDDS, a novel framework that disentangles molecular graphs into causal and spurious substructures, enabling more accurate and interpretable prediction of synergistic drug combinations. Unlike traditional models, CausalDDS utilizes a conditional intervention module and employs noise masking to suppress spurious features, enabling the model to focus on genuinely causal molecular representations. To further refine the learning process, we design optimization objectives grounded in the principles of sufficiency and independence, which ensure that causal substructures are both sufficient to predict synergy outcomes and independent from spurious signals. This principled loss design not only strengthens causal feature attribution but also improves model robustness in scenarios under cold-start or out-of-distribution data. 

The experimental results on two public datasets demonstrate that CausalDDS consistently outperforms state-of-the-art methods across diverse experimental settings. In particular, CausalDDS demonstrates remarkable superiority in challenging scenarios such as cold-start prediction and out-of-distribution generalization. These settings are especially important in drug discovery, where new drugs or novel disease contexts often lack sufficient prior data, and predictive models must therefore extrapolate beyond the training distribution. The strong performance of CausalDDS in these scenarios highlights its robustness and practical value, underscoring its potential for guiding drug repurposing and the design of novel synergistic drug combinations in real-world applications.

Although our framework integrates gene expression profiles of cell lines to capture cellular context, it cannot identify the specific genes or target proteins through which a drug pair achieves its synergistic effect. CausalDDS is designed to predict whether a given drug pair will exhibit synergy in a particular cell line, leveraging transcriptomic features to capture the cellular context. However, it does not trace the predicted synergy back to molecular mechanisms, such as the compensatory pathway or key gene. Without the ability to pinpoint key genes or protein targets modulated by a drug pair in a specific cellular context, the model cannot provide actionable biological insights or guide downstream validation. Future versions of the model could incorporate drug–target interaction data or infer pathway-level effects, enabling a shift from predictive modeling toward mechanism-informed discovery of drug synergy.

Additionally, the current framework is specifically designed for pairwise drug combinations. However, many clinical regimens involve three or more drugs, and the mechanistic interactions in these multi-drug settings are often non-additive and highly context-dependent. Extending the causal disentangle framework to higher-order combinations remains an open challenge but is essential for more realistic modeling of therapeutic strategies.

Overall, the CausalDDS model represents a powerful addition to the drug discovery toolkit, making significant contributions to accelerating the drug development process and improving cost-effectiveness. Its strong performance, interpretability, and generalization ability highlight its practical potential for applications in the biomedical domain.

\section{Methods}

\subsection{The workflow of CausalDDS.} The overall framework of the CausalDDS is shown in Fig. 1, which consists of five modules, namely the Input module, the Molecular Interaction module, the Causal Disentangle module, the Conditional Causal Intervention Module, and the Prediction Module.

\subsection{Input module}
The simplified molecular-input line-entry system (SMILES)\cite{weininger1988smiles} of the drugs is obtained from the DrugBank\cite{wishart2018drugbank} database. We use the open-source chemical informatics software RDKit\cite{landrum2006rdkit} to convert the SMILES string into its 2D molecular graph, where the nodes are atoms and the edges are chemical bonds. Each node is represented as a 133-dimensional vector describing eight pieces of information: the atom atomic number, the atom degree, the formal charge, the atom chirality, the number of hydrogens, the atom hybridization, the atom mass, and whether the atom is aromatic. Each edge is represented as a 14-dimensional vector describing five pieces of information: the bond type, the bond conjugation, the bond stereo, whether the bond is a ring membership, and whether the bond exists. 

For cell lines, we obtain the gene expression data from the Cancer Cell Line Encyclopedia (CCLE)\cite{barretina2012cancer}, which is an independent project that makes an effort to characterize genomes, messenger RNA expression, and anticancer drug dose responses across cancer cell lines. The expression data is normalized through Transcripts Per Million based on the genome-wide read counts matrix. Here, we consider only data related to 651 genes from the COSMIC Cancer Gene Census\cite{forbes2015cosmic}. Considering the intersection of 651 genes and the CCLE gene expression profiles, we select 640 genes from raw expression profiles as input to the model. 

\subsection{Molecular Interaction module} We use a three-layer graph isomorphism network (GIN) block\cite{xu2018powerful} to effectively learn the graph representation on the drug molecular graph. This block updates the atomic feature vectors by aggregating the feature vectors of their corresponding neighboring atoms, connected by chemical bonds, as well as the edge feature vectors. Specifically, the iteration process can be defined as below:
\begin{equation}
\mathbf{h}_i^{(l+1)} = \text{ReLU} \left( \text{GINEConv}^{(l)}\left( \mathbf{h}_i^{(l)}, \mathcal{N}(i), \mathbf{e}_{ij} \right) \right),
\end{equation}
where $\mathbf{h}_i^l$ represents the feature vector of atom $i$ at layer $l$, $\mathcal{N}(i)$ is the set of neighbors of drug $i$, and $\mathbf{e}_{ij}$ denotes the feature vector of the edge between nodes $i$ and $j$. Here, we adopt the Graph Isomorphism Network with edge features (GINEConv) as the message passing operator, which extends the GIN by incorporating edge attributes into the aggregation function. This design allows the model to capture not only atom-level features but also bond-level information, making it particularly suitable for molecular graphs. Given a pair of drugs $\mathbf{D}_1$ and $\mathbf{D}_2$, we first obtain the atom representation matrix: $\mathbf{X}_1 \in \mathbb{R}^{{N_1}\times d}$ and $\mathbf{X}_2 \in \mathbb{R}^{{N_2}\times d}$, respectively, and $N_1$ and $N_2$ denote the number of atoms in the drug $\mathbf{D}_1$ and $\mathbf{D}_2$, respectively.

After acquiring the node representations $\mathbf{X}_1$ and $\mathbf{X}_2$ for the two drugs, we model the nodewise interaction between $\mathbf{D}_1$ and $\mathbf{D}_2$ using an interaction map $\mathbf{I}\in\mathbb{R}^{N_1\times N_2}$, which is defined as follows:
\begin{equation}
   \mathbf{I} = \text{cos}(\mathbf{X}_1, \mathbf{X}_2),
\end{equation}
where $\text{cos}(\cdot, \cdot)$ is the cosine similarity. Using the interaction map $\mathbf{I}$, we calculate the embedding matrices $\mathbf{E}_1$ and $\mathbf{E}_2$ for each drug $\mathbf{D}_1$ and $\mathbf{D}_2$ as follows:
\begin{equation}
    \mathbf{E}_1 = \mathbf{I} \cdot \mathbf{X}_2 , \quad \mathbf{E}_2 = \mathbf{I} \cdot \mathbf{X}_1.
\end{equation}
Subsequently, we construct the final node representation matrix $\mathbf{H}_1$ for drug $\mathbf{D}_1$ by concatenating $\mathbf{X}_1$ and $\mathbf{E}_1$:
\begin{equation}
    \mathbf{H}_1 = (\mathbf{X}_1 \| \mathbf{E}_1).
\end{equation}
We also generate the final node representation matrix $\mathbf{H}_2$ for drug $\mathbf{D}_2$ similarly.

\subsection{Causal Disentangle module}
Identifying causal and spurious substructures in molecular graphs is crucial to improve the generalizability of the model and the interpretability of its mechanisms. Since the causal substructure of a molecule is influenced by another drug in the reaction, we propose a model that separates the causal substructure $\mathbf{C}_1$ and the spurious substructure $\mathbf{S}_1$ in the representation space of $\mathbf{D}_1$ by masking the node representation matrix $\mathbf{H}_1$, which contains information about both drug $\mathbf{D}_1$ and $\mathbf{D}_2$. Specifically, given an representation of node $i$ denoted as $\mathbf{H}_1^i$, we parametrize the importance $p_i$ of node $i$ using a multilayer perception (MLP) as follows: 
\begin{equation}
    p_i = \text{MLP}(\mathbf{H}_1^i).
\end{equation}
Using the calculated importance $p_i$ of node $i$, we obtain the representation of node $i$ in the causal substructure $\mathbf{C}_1^i$ and the spurious substructure $\mathbf{S}_1^i$ by applying a mask $\epsilon$ sampled from a normal distribution as follows: 
\begin{equation}
    \mathbf{C}_1^i = \lambda_i \mathbf{H}_1^i + (1 - \lambda_i) \epsilon, \quad \mathbf{S}_1^i = (1 - \lambda_i) \mathbf{H}_1^i,
\end{equation}
where $\lambda_i \sim \text{Bernoulli}(p_i)$ and $\epsilon \sim \mathcal{N}(\mu_{\mathbf{H_1}}, \sigma_{\mathbf{H_1}}^2)$. Note that
$\mu_{\mathbf{H}_1}$ and $\sigma^2_{\mathbf{H}_1}$ represent the mean and variance of the node embeddings in $\mathbf{H}_1$, respectively.
In doing so, our objective is to automatically learn the causal substructure that is causally related to the target variable $Y$, while effectively suppressing the spurious substructure. Subsequently, we adopt the Gumbel-sigmoid method to sample $\lambda_i$ as follows:
\begin{equation}
    \lambda_i = \text{Sigmoid} \left( \frac{1}{t} \left( \log\frac{p_i}{1 - p_i} + \log\frac{u}{1 - u} \right) \right),
\end{equation}
where $u \sim \text{Uniform(0,1)}$, and $t$ is the temperature hyperparameter. To construct substructure-level feature vectors, we use the Set2Set read-out function\cite{vinyals2015order} to aggregate the substructure feature from the learned causal substructure representation matrix $\mathbf{C}_1$ and the spurious substructure representation matrix $\mathbf{S}_1$. As a result, we obtain the substructure level representation $\mathbf{Z}_{\mathbf{C}_1}$ and $\mathbf{Z}_{\mathbf{S}_1}$ of drug $\mathbf{D}_2$. Similarly, we also obtain the substructure-level representation $\mathbf{Z}_{\mathbf{C}_2}$ and $\mathbf{Z}_{\mathbf{S}_2}$ of drug $\mathbf{D}_1$.

\subsection{Conditional Causal Intervention Module}
To mitigate spurious correlations and encourage the model to focus on causal substructures, we introduce a conditional causal intervention mechanism. Given a batch of molecular samples, we randomly permute their indices to construct a counterfactual input distribution under the intervention condition. Specifically, given a drug pair $\mathbf{D}_1$ and $\mathbf{D}_2$, we extract the spurious substructure $\tilde{\mathbf{S}}_1$ and $\tilde{\mathbf{S}}_2$ by modeling the interactions between $\mathbf{D}_1$ and the set of other drugs $D$ in the training data and also for $\mathbf{D}_2$. We then apply global mean pooling to extract substructure-level representations $\mathbf{Z}_{\tilde{\mathbf{S}}_1}$ and $\mathbf{Z}_{\tilde{\mathbf{S}}_2}$ from $\tilde{\mathbf{S}}_1$ and $\tilde{\mathbf{S}}_2$, respectively. Based on these representations, we optimize the loss of the conditional causal intervention module:
\begin{equation}
    \mathcal{L}_{\text{inter}} = \sum_{(\mathbf{D}_1, \mathbf{D}_2) \in \mathcal{D}} \sum_{\tilde{\mathbf{S}}_1,\tilde{\mathbf{S}}_2} \mathcal{L}(Y, \mathbf{z}_{\mathbf{C}_1}, \mathbf{z}_{\mathbf{C}_2}, \mathbf{z}_{\tilde{S}_1}, \mathbf{z}_{\tilde{S}_2}),
\end{equation}
where loss function $\mathcal{L}$ can be modeled as the cross entropy loss for classification and the root mean squared error loss for regression.

\subsection{Prediction Module}
The causal substructure representation $\mathbf{Z}_{C_1}$ of drug $\mathbf{D}_1$, the causal substructure representation $\mathbf{Z}_{C_2}$ of drug $\mathbf{D}_2$, and the expression feature $\mathbf{Z}_{cell}$ of cell line are concatenated and fed into an MLP network for drug synergy prediction. For the binary classification task, the last layer of MLP outputs the probability $\hat{y}$, indicating the probability that the drug pair is a synergistic drug combination in the cell line. For the regression task, the last layer of MLP outputs the synergy score $\hat{y}$, which measures the therapeutic effect of the drug combination in the given cell line.

\subsection{Training Objective}
The training objective is formulated according to two fundamental principles: sufficiency and independence. The principle of sufficiency implies that the causal substructure should retain all critical information relevant to the prediction of the label $y$. The loss of sufficiency is formulated in the classification task as:
\begin{equation}
\hat{y}_i = \text{MLP} \left( \mathbf{z}_{C_1}^{(i)} \,\|\, \mathbf{z}_{C_2}^{(i)} \,\|\, \mathbf{z}_{\text{cell}}^{(i)} \right),
\end{equation}
\begin{equation}
\mathcal{L}_{\text{suf}} = -\frac{1}{N} \sum_{i=1}^{N} \left[ y_i \cdot \log(\hat{y}_i) + (1 - y_i) \cdot \log(1 - \hat{y}_i) \right],
\end{equation}
where $\mathbf{z}_{C_1}^{(i)}$ and $\mathbf{z}_{C_2}^{(i)}$ denote the representations of causal substructure of the two drugs in the $i$-th sample, and $\mathbf{z}_{\text{cell}}^{(i)}$ represents the embedding of cell line in the $i$-th sample. The loss of sufficiency is formulated in the regression task as:

\begin{equation}
\mathcal{L}_{\text{suf}} = \frac{1}{N} \sum_{i=1}^{N} \left( y_i - \hat{y}_i \right)^2
\end{equation}

The principle of independence states that the spurious substructure should not contain any information relevant to the prediction of the label $y$. The loss of independence is formulated as:
\begin{equation}
\hat{y}_{ind} = \text{MLP} \left( \mathbf{z}_{S_1}^{(i)} \,\|\, \mathbf{z}_{S_2}^{(i)} \,\|\, \mathbf{z}_{\text{cell}}^{(i)} \right),
\end{equation}
\begin{equation}
\mathcal{L}_{\text{ind}} = \text{KL}(\hat{y}_{\text{ind}} \,\|\, \mathcal{U}),
\end{equation}
where $\mathbf{z}_{S_1}^{(i)}$ and $\mathbf{z}_{S_2}^{(i)}$ denote the spurious substructure representations of the two drugs in the $i$-th sample, and $\mathcal{U}$ is the uniform distribution over two classes. Finally, we train the model with the following objective:
\begin{equation}
    \mathcal{L}_{\text{tol}} = \mathcal{L}_{\text{suf}} + \mathcal{L}_{\text{ind}} + \mathcal{L}_{\text{inter}}.
\end{equation}

\subsection{Evaluation protocols}
Six different experimental settings are proposed to evaluate the generalization performance in drug synergy prediction, namely warm start, unseen drug pair, unseen cell line, unseen drug, scaffold split, and SIMPD split.

\textbf{Warm start} Five-fold cross-validation is implemented by randomly dividing all samples into five equal folds. The objective in this scenario was to rediscover known anti-cancer drug synergies.

\textbf{Unseen drug pair} Five-fold cross-validation is implemented by randomly splitting the samples at the cell line level to guarantee the test set only contains unseen drug pairs that are not included in the training set.

\textbf{Unseen cell line} Five-fold cross-validation is conducted by splitting the samples at the drug combination level to ensure the test set only contains unseen cell lines that are not included in the training set.

\textbf{Unseen drug} Five-fold cross-validation is conducted by splitting the samples at the drug level to ensure the test set only contains unseen drugs that are not included in the training set.

\textbf{Scaffold split} Five-fold cross-validation is conducted by splitting the samples at the drug structural framework level to ensure the test set only contains unseen drug scaffolds that are not included in the training set\cite{fang2022geometry}.

\textbf{SIMPD split} Five-fold cross-validation is conducted by splitting the samples at the drug temporal information level to ensure that the temporal information of the drugs in the test set does not appear in the training set\cite{landrum2023simpd}.

\subsection{Datasets and metrics}
We evaluate CausalDDS alongside state-of-the-art baselines using two publicly available drug synergy datasets: DrugCombDB\cite{liu2020drugcombdb} and ONEIL-COSMIC\cite{o2016unbiased}. Details of the datasets and metrics used in this study are provided in Supplementary Note 2 and Supplementary Table 10. Both data sets provide drug SMILES sequences, cell line gene expression data, and experimentally measured drug synergy scores. We evaluate the performance of different models on the drug synergy prediction task using eight distinct metrics: four for regression and four for classification. For the classification task, we consider four metrics: the area under the receiver operating characteristic (AUC), the area under the precision-recall curve (AUPR), Cohen’s Kappa (KAPPA), and F1 Score (F1). For the regression task, we consider four metrics: the Root Mean Square Error (RMSE), the Mean Absolute Error (MAE), the Pearson correlation coefficient (PCC), and R-Squared (R\textsuperscript{2}). We report the mean and standard deviation of these metrics using five-fold cross-validation for each dataset.

\section{Data availability}
The data used in this study are publicly available and can be accessed at [https://github.com/Royluoyi123/CausalDDS]/
\section{Code availability}
The source code and implementation details for this study are freely available via [https://github.com/Royluoyi123/CausalDDS].

{\small
\bibliographystyle{unsrt}
\bibliography{egbib}
}

\section{Acknowledgements}
This work was also carried out in part using computing resources at the High Performance Computing Center of Central South University.

\section{Funding}
This work was supported in part by the National Natural Science Foundation of China (No. 62350004, No. 62332020), the Project of Xiangjiang Laboratory (No. 23XJ01011). 

\section{Author contributions}
J.X.W. conceived and designed this project. Y.L., H.C.Z., and J.X.W. conceived, designed, and implemented the method. Y.L., X.L., Y.Y.Z, X.Y.L, and Y.W.L. conducted the analyzes and prepared the figures. Y.L. and H.C.Z. wrote the paper. J.X.W. revised and proofread the manuscript. All authors have read and approved the final version of this paper.

\section{Conflict of Interest}
The authors declare no competing interests.

\end{document}